\documentclass[submission,copyright,creativecommons]{eptcs}

\providecommand{\event}{ICLP 2026} \usepackage{iftex}
\ifpdf
\usepackage{underscore}
\usepackage[T1]{fontenc}
\else
\usepackage{breakurl}
\fi

\usepackage{xspace}
\usepackage{xparse}
\usepackage{environ}
\usepackage{algorithm}
\usepackage{algpseudocode}
\usepackage{bm}
\usepackage[english]{babel}
\usepackage{amsthm}
\usepackage{amsmath}
\usepackage{amssymb}
\usepackage{mathtools}
\usepackage{graphicx}
\usepackage{hyperref}
\usepackage{xfrac}
\usepackage{stackengine}
\usepackage{float}
\usepackage{tikz}
\usepackage{xcolor}
\usepackage{parskip}

\def\thm@space@setup{
  \thm@preskip=0pt \thm@postskip=0pt }
\newtheoremstyle{customspace}{0pt}{0pt}{}{}{\bfseries}{.}{.5em}{}
\theoremstyle{remark}
\theoremstyle{customspace}

\def\thm@space@setup{
  \thm@preskip=0pt
  \thm@postskip=0pt
}

\newtheorem{decisionproblem}{Decision Problem}
\newtheorem{theorem}{Theorem}
\newtheorem{corollary}{Corollary}
\newtheorem{lemma}{Lemma}
\newtheorem{proposition}{Proposition}
\newtheorem{definition}{Definition}
\newtheorem{example}{Example}
\newtheorem{remark}{Remark}

\usepackage[createShortEnv, conf={proof at the end, restate}]{proof-at-the-end}

\newcommand{\ltePrecision}{\subseteq_p^2}

\newcommand{\define}{\coloneqq}
\newcommand{\union}{\mathbin{\cup}}
\newcommand{\unionBig}{\bigcup}

\newcommand{\intersect}{\mathbin{\cap}}
\newcommand{\intersectBig}{\bigcap}

\newcommand{\lfp}{\textbf{lfp}}

\let\restrictionWithSpaces\restriction
\renewcommand{\restriction}{{\restrictionWithSpaces}}

\newcommand{\Prog}{\mathcal{P}}

\newcommand{\body}{body}
\newcommand{\bodyp}{body^{+}}
\newcommand{\bodyn}{body^{-}}
\newcommand{\head}{head}

\newcommand{\Not}{\bm{not}}

\newcommand{\Ont}{\mathcal{O}}

\newcommand{\unionPlus}{\mathbin{\ensurestackMath{\stackinset{c}{0pt}{c}{1pt}{\scriptscriptstyle{\bm{+}}}{\union}}} }
\newcommand{\unionMinus}{\mathbin{\ensurestackMath{\stackinset{c}{-0.pt}{c}{0pt}{\text{-}}{\union}}} }
\newcommand{\unionTop}{\mathbin{\sim}}
\newcommand{\intersectMinus}{\mathbin{\ensurestackMath{\stackinset{c}{-0.0pt}{c}{0pt}{\text{-}}{\intersect}}} }

\newcommand{\dlModels}{\models_{{\scriptscriptstyle {\sf DL}}}}
\newcommand{\dlModelsMin}{\models_{{\scriptscriptstyle {\sf {\wedge}DL}}}}

\newcommand{\alignedreduct}[2]{{#1}_{bn.}^{#2}}

 \newcommand{\definitionBody}[2]{\hypertarget{glossary:#1}{#2}}
\newcommand{\definitionLink}[2]{#2\xspace}
\newcommand{\Definition}[1]{\definitionLink{#1}{Definition~\ref{definition:#1}}}
\NewEnviron{definitionOf}[1]{
    \begin{definition}\label{definition:#1}\label{glossary:#1}
        \definitionBody{#1}{\BODY}
    \end{definition}
}
\NewEnviron{definitionOfWtitle}[2]{
    \begin{definition}[#2]\label{definition:#1}\label{glossary:#1}
        \definitionBody{#1}{\BODY}
    \end{definition}
}
\NewEnviron{propositionOf}[1]{
    \begin{proposition}\label{proposition:#1}
        \BODY
    \end{proposition}
}
\NewEnviron{theoremOf}[1]{
    \begin{theorem}\label{theorem:#1}
        \BODY
    \end{theorem}
}
\newcommand{\Proposition}[1]{\hyperref[proposition:#1]{Proposition~\ref{proposition:#1}}}
\NewEnviron{exampleOf}[1]{
    \begin{example}\label{example:#1}
        \BODY
    \end{example}
}
\newcommand{\Example}[1]{\definitionLink{#1}{Example~\ref{example:#1}}}
\NewEnviron{lemmaOf}[1]{
    \begin{lemma}\label{lemma:#1}
        \BODY
    \end{lemma}
}
\newcommand{\Lemma}[1]{\hyperref[lemma:#1]{Lemma~\ref{lemma:#1}}}
\NewEnviron{remarkOf}[1]{
    \begin{remark}\label{remark:#1}
        \BODY
    \end{remark}
}
\newcommand{\Corollary}[1]{\hyperref[corollary:#1]{Corollary~\ref{corollary:#1}}}
\NewEnviron{corollaryOf}[1]{
    \begin{corollary}\label{corollary:#1}
        \BODY
    \end{corollary}
}
\newcommand{\DecisionProblem}[1]{\hyperref[decisionproblem:#1]{Decision Problem~\ref{decisionproblem:#1}}}
\NewEnviron{decisionproblemOf}[1]{
    \begin{decisionproblem}\label{decisionproblem:#1}
        \BODY
    \end{decisionproblem}
}
\NewEnviron{decisionproblemOfWtitle}[2]{
    \begin{decisionproblem}[#2]\label{decisionproblem:#1}
        \definitionBody{#1}{\BODY}
    \end{decisionproblem}
}

\NewEnviron{propositionOfE}[1]{
    \begin{propositionE}\label{proposition:#1}
        \BODY
    \end{propositionE}
}
\NewEnviron{theoremOfE}[1]{
    \begin{theoremE}\label{theorem:#1}
        \BODY
    \end{theoremE}
}
\NewEnviron{lemmaOfE}[1]{
    \begin{lemmaE}\label{lemma:#1}
        \BODY
    \end{lemmaE}
}
\NewEnviron{corollaryOfE}[1]{
    \begin{corollaryE}\label{corollary:#1}
        \BODY
    \end{corollaryE}
}

\newcommand{\Theorem}[1]{\hyperref[theorem:#1]{Theorem~\ref{theorem:#1}}}

\newcommand{\Section}[1]{Section~\ref{section:#1}}

\newcommand{\Antimonotone}{\definitionLink{monotone}{antimonotone}}

\newcommand{\Monotone}{\definitionLink{monotone}{monotone}}

\newcommand{\Prefixpoint}{\definitionLink{deterministicprefixpoint}{prefixpoint}}

{}

{}

{}

{}

\let\topNoLink\top{}
\renewcommand{\top}{\definitionLink{topbot}{\topNoLink}}
\let\botNoLink\bot{}
\renewcommand{\bot}{\definitionLink{topbot}{\botNoLink}}

{}

{}

{}

{}

{}

{}

\newcommand{\AnInterpretation}{\definitionLink{interpretation}{an interpretation}}
\newcommand{\AnInterpretationO}{\definitionLink{interpretation}{An interpretation}}
\newcommand{\Interpretation}{\definitionLink{interpretation}{interpretation}}

\newcommand{\Interpretations}{\definitionLink{interpretation}{interpretations}}

\let\restrictionNoLink\restriction
\renewcommand{\restriction}{\definitionLink{restriction}{\restrictionNoLink}}

\newcommand{\AlignedReduct}{\definitionLink{alignedreduct}{bound reduct}}

\newcommand{\AlignedPositive}{\definitionLink{propositionaldl}{aligned-positive}}
\newcommand{\AlignedAnswerSet}{\definitionLink{alignedanswerset}{bound answer set}}
\newcommand{\AnAlignedAnswerSet}{\definitionLink{alignedanswerset}{a bound answer set}}
\let\alignedreductNoLink\alignedreduct
\renewcommand{\alignedreduct}[2]{\definitionLink{alignedreduct}{\alignedreductNoLink{#1}{#2}}}

\newcommand{\Model}{\definitionLink{interpretationmodel}{model}}

\newcommand{\AnAnswerSetO}{\definitionLink{stablemodels}{An answer set}}

\let\NotNoLink\Not
\renewcommand{\Not}{\definitionLink{notop}{\NotNoLink}}

\let\ltePrecisionNoLink\ltePrecision
\renewcommand{\ltePrecision}{\definitionLink{lteprecision}{\ltePrecisionNoLink}}

\newcommand{\DLProgram}{\definitionLink{dlprogram}{DL program}}
\newcommand{\DLPrograms}{\definitionLink{dlprogram}{DL programs}}

\newcommand{\ADLProgram}{\definitionLink{dlprogram}{a DL program}}
\newcommand{\ADLProgramO}{\definitionLink{dlprogram}{A DL program}}
\newcommand{\StrongWellSupportedModel}{\definitionLink{strongwellsupportedmodel}{well-supported answer set}}

\newcommand{\WellSupportedSemantics}{\definitionLink{strongwellsupportedsemantics}{well-supported}}

\newcommand{\AlignedDLAtom}{\definitionLink{dlatomaligned}{aligned}}
\newcommand{\AnAlignedDLAtom}{\definitionLink{dlatomaligned}{an aligned}}

\newcommand{\AlignedDLAtomO}{\definitionLink{dlatomaligned}{Aligned}}
\newcommand{\AlignedModel}{\definitionLink{alignedmodel}{bound}}
\newcommand{\AnAlignedModel}{\definitionLink{alignedmodel}{a bound}}

\newcommand{\UnalignedDLAtom}{\definitionLink{dlatomaligned}{unaligned}}
\newcommand{\AnUnalignedDLAtom}{\definitionLink{dlatomaligned}{an unaligned}}

\newcommand{\DLAtom}{\definitionLink{dlatom}{DL-atom}}
\newcommand{\DLAtoms}{\definitionLink{dlatom}{DL-atoms}}
\newcommand{\DLAtomsO}{\definitionLink{dlatom}{DL-atoms}}
\newcommand{\ADLAtom}{\definitionLink{dlatom}{a DL-atom}}
\newcommand{\ADLAtomO}{\definitionLink{dlatom}{A DL-atom}}

\let\dlModelsNoLink\dlModels
\renewcommand{\dlModels}{\definitionLink{dlmodels}{\dlModelsNoLink}}
\let\dlModelsMinNoLink\dlModelsMin
\renewcommand{\dlModelsMin}{\definitionLink{dlmodelsmin}{\dlModelsMinNoLink}}

\title{A New Well-Supported Semantics \\ for Description Logic Programs}
\author{
Spencer Killen \qquad\qquad Jia-Huai You
\institute{University of Alberta\\
Alberta, Canada}
\email{\quad sjkillen@ualberta.ca \quad\qquad jyou@ualberta.ca}
}
\def\titlerunning{A New Well-Supported Semantics for Description Logic Programs}
\def\authorrunning{S. Killen \& J You}
\begin{document}
\maketitle
\begin{abstract}
Description logic programs are a powerful formalism for combining rules with ontologies.
The well-supported semantics for description logic programs ensures that no answer sets rely on cyclic dependencies.
Most popular semantics for logic programming have this property of well-supportedness.
We recognize two limitations of the current well-supported semantics for DL programs: its increased computational complexity for the consistency problem and its lack of a reduct transformation characterization.

In this work, we present a new semantics which evaluates ontological atoms more strictly than the current semantics.
This keeps the complexity of its consistency problem NP-complete, rather than increasing it to the second level of the polynomial hierarchy.
Additionally, we identify a syntactic class of description logic programs for which our new semantics is equivalent to the current semantics.
We characterize our semantics using a fixpoint operator and a reduct-based transformation.
Our new semantics is a strict subset of the current well-supported semantics, so it maintains the prior notion  of well-supportedness while inducing its own stricter notion.
We prefer our new notion of well-supportedness due to its similarities with logic programming.
\end{abstract}
\section{Introduction}
DL programs (description logic programs) ~\cite{dlpprograms} equip logic programs with ontological queries via description logics.
This enables mixed reasoning under the closed- and open-world assumptions.
\DLPrograms have garnered a wide amount of attention since their introduction.
The approach is loose as opposed to the tight reasoning of hybrid MKNF~\cite{motikreconciling2010}.
In \DLPrograms, the truth value of a program atom does not inherently affect the ontology.
Reasoning occurs unidirectionally by embedding queries in a program.
A query embedded in a logic program communicates with the ontology in isolation from all other queries.
For example, $p \leftarrow DL[; S](t)$ is \ADLProgram comprised of a single rule that queries the ontology about $S(t)$ using the \DLAtom $DL[; S](t)$.
If the ontology accompanying the program above is $\{ S(t) \}$, then the query succeeds; if it is $\emptyset$, then the \DLAtom is false.
\DLAtomsO, which are only permitted in the body of rules, are true in the logic program if their query is true.
Notably, each \DLAtom acts as an isolated query, and queries do not affect each other.
\ADLAtomO can also specify input atoms to bind to the ontology.
Using this mechanism, one can make concepts classically false in the description logic with negation as failure as the source of
this falsity.
For instance, the program $h \leftarrow DL[S \intersectMinus p; \neg S](t)$ binds the logic program predicate ``$p$'' to the description logic object $\neg S(t)$ using negation as failure
($\intersectMinus$).
That is, if ``p(t)'' is false, then $\neg S(t)$ is temporarily made to be true in the ontology, and the query $\neg S(t)$ succeeds.

Shen~\cite{dlwellsupported} demonstrates that DL answer sets under Eiter et al.'s original semantics~\cite{dlpprograms} may not be well-supported.
Well-supportedness is a desired property for semantics because it guarantees that the truth of an atom can be derived
independently without relying on itself~\cite{fagesws}.
A well-supported model has no cycles in the derivation of atoms, that is, the truth of an atom $p$ cannot depend on an atom which is in turn dependent on $p$.
Shen\ identifies a subset of Eiter et al.'s answer sets to define a well-supported semantics.
This semantics is faithful to Eiter et al.'s original ``strong'' semantics in that every well-supported answer set is a strong answer
set under Eiter et al.'s semantics.

For Shen's semantics, the complexity of determining whether a model exists is $\Sigma_2^P$-complete~\cite{hexaft} while for the original semantics, this problem is NP-complete.\footnote{We impose some restrictions on ontologies for complexity analysis purposes in \Section{preliminaries}.}
Such an increase in complexity complicates the construction of a solver.
If the complexity were in the first level (NP-complete), it would require a less sophisticated solver.
To ensure well-supportedness under Shen's semantics, an exponentially large range of \Interpretations must be evaluated for each \DLAtom query.
This raises a question as to whether we can isolate this complexity.
For disjunctive logic programs, it is known that head-cycles~\cite{disprop} cause their increased complexity~\cite{disjunctive_complexity} compared to nondisjunctive programs.
A question arises as to whether there is a syntactic class of \DLPrograms whose complexity remains in the first level of the polynomial hierarchy under the well-supported semantics.

In this work, we define a new well-supported semantics which further refines Shen's well-supported semantics.
Our main contributions are as follows: we
    (1) introduce a new well-supported semantics, characterized by fixpoints of an operator,
    (2) argue that the new semantics is NP-complete,
    (3) demonstrate that our semantics is a strict subset of the current well-supported semantics,
    (4) introduce an equivalent reduct transformation characterization of our new semantics, and
    (5) identify a syntactic class of \DLPrograms where our semantics is equivalent to the well-supported semantics.

Because our new semantics is stricter than the well-supported semantics, every new answer set is well-supported.
However, some well-supported answer sets are removed from the new semantics.
Some of these removed answer sets exhibit properties of cyclical support; however, this ultimately depends on the definition of well-supportedness one adopts.
Our new semantics induces an alternative definition of well-supportedness which aligns better with the characteristics of logic programs.

This paper is comprised of several sections of the following topics: an introduction of preliminaries (\ref{section:preliminaries}), some motivation and formulation of our new semantics (\ref{section:why}), a full fixpoint characterization of our semantics and related results~(\ref{section:new}), an alternative reduct-based characterization of our new semantics (\ref{section:reduct}), and finally a discussion of further implications of our results and related work (\ref{section:discussion}).
 % Select proofs are included in \Section{proofs}.

\section{Preliminaries}\label{section:preliminaries}
Following~\cite{dlwellsupported},
an ontology (a.k.a.\ a DL knowledge base)~\cite{descriptionlogic} $\Ont$ is a finite set of axioms constructed using a vocabulary $\Sigma_{\Ont} = ({\bf A} \union {\bf R}, {\bf I})$ of mutually disjoint countable sets: {\bf A}  contains atomic concepts, {\bf R} contains atomic roles, and {\bf I} contains individuals.
Because description logics are decidable fragments of first-order logic, an ontology $\Ont$ has first-order semantics.
For complexity analysis purposes, we assume the entailment relation of an ontology can be computed in polynomial time and that the vocabulary $\Sigma_{\Ont}$ is finite.
Let {\bf P} be a set of predicate symbols such that there is no overlap with {\bf I}, {\bf A}, or {\bf R}.
Let {\bf C} be a set of constants from {\bf I}.
A {\em term} is either a constant from {\bf C} or a variable.
An {\em atom} is of the form $p(t_1,\dots t_m)$ where $p$ is a predicate from {\bf P} and each $t_i$ is a term.
The {\em Herbrand base} consists of all such atoms that are variable-free.
An {\em \Interpretation} is a subset of the Herbrand base which treats contained elements as true and all other elements as false.
A range of \Interpretations $(E, I)$ is a pair of \Interpretations s.t.\ $E \subseteq I$.
Intuitively, a range contains every \Interpretation $F$ s.t.\ $E \subseteq F \subseteq I$.
An equality ($p = q$) and inequality ($p \not= q$) relate two terms $p$ and $q$.
A {\em DL-query} $Q(\bm{t})$ takes $Q$ to be an equality/inequality symbol, a concept, a role, a concept inclusion axiom, or their negation constructed from ${\bf A} \union {\bf R}$.
A {\em DL-atom} is of the form $DL[S_1 \mathbin{\textrm{op}_1} p_1, \dots, S_m \mathbin{\textrm{op}}_m p_m; Q](\bm{t})$ where each $S_i$ is a concept or role constructed from ${\bf A} \union {\bf R}$, each $\textrm{op}_i$ is one of $\unionPlus$, $\unionMinus$, or $\intersectMinus$, and $p_i \in P$ is a unary predicate symbol if $S_i$ is a concept and a binary predicate symbol otherwise, and $Q(\bm{t})$ is a DL-query.
We call $\unionPlus$ and $\unionMinus$ the $\union$-operators and $\intersectMinus$ the
$\intersect$-operator.
Likewise, $\union$-\DLAtoms only contain $\union$-operators while $\intersect$-\DLAtoms
only contain $\intersect$-operators.

\definitionBody{dlprogram}{
\ADLProgramO $\Prog$ is a set of rules accompanied by an ontology $\Ont$.
Each rule $r$ has a head component $\head(r)$, a positive body $\bodyp(r)$, and a negative body $\bodyn(r)$.
The head is a single atom, and the bodies are both sets containing atoms and \DLAtoms.
We write a rule $r$ where $\head(r) = h$, $\bodyp(r) = \{ b_1, \dots, b_n \}$, and $\bodyn(r) = \{ c_1, \dots, c_m \}$ as
$h \leftarrow b_1, \dots, b_n, \Not~c_1, \dots, \Not~c_m$.
We use $\body(r)$ to denote the set $\{ b_1, \dots, b_n, \Not~c_1, \dots, \Not~c_m\}$.
A \DLAtom is positive if it is in $\bodyp(r)$.
We call $\Not~\phi$ a negative \DLAtom where $\phi$ is a \DLAtom.
We assume that programs are ground, that is, they do not contain any variables.
The ontology $\Ont$ accompanying a \DLProgram $\Prog$ is a part of $\Prog$, and as such, we do not reference it explictly unless we need too.
In our examples of concrete \DLPrograms, we assume $\Ont = \emptyset$.
}
A \DLProgram without \DLAtoms has the answer set semantics~\cite{answersetsemantics}.

It is convenient for us to use $\in$ to relate \DLAtoms with their substrings.
For example, given $\phi$ to be the \DLAtom $DL[S \unionPlus p, S \intersectMinus p; S](t)$, we have $\unionPlus~ p \in\phi$ and $S~\unionPlus \in \phi$.
Next, we define how the operators in a \DLAtom bind atoms from the logic program to concepts and roles in the ontology.
\begin{definition}
Given \ADLAtom $\phi$ and \AnInterpretation $I$, we define $query^{\union}_{\phi}(I)$ and $query^{\intersect}_{\phi}(I)$~\footnotetext{Following Wang et al.~\cite{eliminatingnonmonotonicdlatoms}, we limit $p(\bm{e})$ to be in the Herbrand base for negation safety.} to be the smallest sets such that
\vspace{-.25cm}
\begin{align*}
S(\bm{e}) \in query^{\union}_{\phi}(I) &\textrm{ if } S \unionPlus p \in \phi \textrm{ and } p(\bm{e}) \in I \\
\neg S(\bm{e}) \in query^{\union}_{\phi}(I) &\textrm{ if } S \unionMinus p \in \phi \textrm{ and } p(\bm{e}) \in I \\
\neg S(\bm{e}) \in query^{\intersect}_{\phi}(I) &\textrm{ if } S \intersectMinus p \in \phi \textrm{ and } p(\bm{e}) \not\in I \textrm{ and } p(\bm{e}) \in HB~\footnotemark[1]
\end{align*}
As a shorthand for the entire query, we define $query_{\phi}(I) \define query^{\union}_{\phi}(I) \union query^{\intersect}_{\phi}(I)$.
\end{definition}
Intuitively, $query_{\phi}(I)$ evaluates the operators inside \ADLAtom given \AnInterpretation $I$.
For example, if $\phi$ is the \DLAtom $DL[S \unionPlus p, R \intersectMinus q; Q](t)$ and $I$ is the \Interpretation $\{ p(t) \}$, we have $query_{\phi}(\{ p(t) \}) = \{ S(t), \neg R(t), \neg R(r) \}$ assuming the Herbrand base contains $q(t)$ and $q(r)$ alone for the predicate $q$.
Given that $p(t)$ is true in the \Interpretation $\{ p(t) \}$, the \DLAtom binds $p$ to $S$ using $\unionPlus$, so the set $query^{\union}_{\phi}(\{
p(t) \})$ contains $S(t)$.
Similarly, because $q(t)$ and $q(r)$ are in the Herbrand base but not true in $I$, the $\intersectMinus$-operator adds the negated concepts s.t.\ $query^{\intersect}_{\phi}(\{ p(t) \}) = \{ \neg R(t), \neg R(r) \}$.
Together, we have $query_{\phi}(\{ p(t) \}) = query^{\union}_{\phi}(\{ p(t) \}) \union query^{\intersect}_{\phi}(\{ p(t) \}) = \{ S(t), \neg R(t), \neg R(r) \}$.
As we will soon see, the result from $query_{\phi}(I)$ is combined with the ontology to evaluate the \DLAtom.
First, we must define the entailment relations for both a single \Interpretation and a range of \Interpretations that Shen uses as a basis to define the well-supported semantics~\cite{dlwellsupported}.
\pagebreak
\begin{definitionOf}{dlmodels}
Let $\phi$ be a \DLAtom $DL[S_1 \mathbin{\textrm{op}_1} p_1, \dots, S_m \mathbin{\textrm{op}}_m p_m; Q](\bm{c})$ inside \ADLProgram $\Prog$ with the ontology $\Ont$, let $p(\bm{t})$
be an atom, and let $r \in \Prog$.
For a single \Interpretation $F$, we have
\begin{align*}
F &\dlModels p(\bm{t}) &\textrm{iff $p(\bm{t}) \in F$} \\
F &\dlModels \Not~ p(\bm{t}) &\textrm{iff $p(\bm{t}) \not\in F$} \\
F &\dlModels \phi &\textrm{iff ${\Ont \union query_{\phi}(F) \models Q}(\bm{c})$} \\
F &\dlModels \Not~\phi &\textrm{iff ${\Ont \union query_{\phi}(F) \not\models Q}(\bm{c})$} \\
F &\dlModels r &\textrm{iff $F \dlModels \body(r)$}\textrm{~implies }F &\dlModels \head(r)
\end{align*}
The relation above is lifted to a range of \Interpretations $(E, I)$ as follows:
\begin{align*}
(E, I) &\dlModels \omega &\textrm{iff $F \dlModels \omega$ for each $F$ s.t.\ $E \subseteq F \subseteq I$.}
\end{align*}
where $\omega$ is any valid form for $F \dlModels \omega $ listed above.
\end{definitionOf}
Returning to our previous example, the \DLAtom  $DL[S \unionPlus p, R \intersectMinus q; Q](t)$, with the \Interpretation $\{ p(t) \}$, is evaluated using $\Ont \union \{ S(t), \neg R(t), \neg R(r)
\} \models Q(t)$.
The relation $(E, I)\dlModels \phi$ checks the \DLAtom $\phi$'s query against all \Interpretations in a given \Interpretation range $(E, I)$ (That is, each $F$ s.t.\ $E \subseteq F \subseteq I$).
Shen\ uses the relation in \Definition{dlmodels} to define an operator over \Interpretation ranges.
\begin{definitionOf}{dlmodelsconsequence}
$\Gamma_{\Prog}(E, I) \define \{ \head(r) ~|~ r \in \Prog, (E, I) \dlModels \body(r) \}$
\end{definitionOf}
With this operator, we can introduce well-supported answer sets.
While Shen~\cite{dlwellsupported} introduces a strong and a weak variant, both of which refine Eiter et al.'s strong DL semantics, we
focus on the strong variant.
We use $\lfp~f(\cdot)$ to denote the {\em $\subseteq$-least fixpoint} of $\lambda x, f(x)$.
\definitionBody{lfp}{\definitionBody{monotone}{ It is well known that a {\em $\subseteq$-\Monotone} function $f$, that is, a
function such that $x \subseteq y$ implies $f(x) \subseteq f(y)$, has a least fixpoint which is its $\subseteq$-least
\Prefixpoint~\cite{tarskilatticetheoretical1955}.}}
\definitionBody{prefixpoint}{ A {\em \Prefixpoint} of $f$ is an element $x$ such that $f(x) \subseteq x$.}
Shen\ demonstrates that $\Gamma_{\Prog}(\cdot, I )$ is $\subseteq$-\Monotone~\cite{dlwellsupported}, thus the following
definition is well-defined.
\begin{definitionOf}{strongwellsupportedmodel}
$I$ is a \StrongWellSupportedModel of a \DLProgram $\Prog$ if $\lfp~\Gamma_{\Prog}(\cdot, I ) = I$.
\end{definitionOf}
Shen~\cite{dlwellsupported} notes that neither Eiter et al.'s~\cite{dlpprograms} weak nor their strong answer set semantics for
\DLPrograms satisfy Fages'~\cite{fagesws} property of well-supportedness.
That is, it is possible to construct an answer set by which atoms are true only because of cyclical support.
Below, we borrow Shen's example~\cite{dlwellsupported}.
\begin{exampleOf}{dlprogramnotws}
Have $\Prog$ be the \DLProgram consisting of the rule
${p(t) \leftarrow DL[S \unionPlus p, S' \intersectMinus q; S \sqcap \neg S'](t)}$.
Eiter et al.'s original semantics~\cite{dlpprograms} is fully reduct-based: Given a \Model $I$ of $\Prog$ ($I \dlModels r$), nonmonotonic \DLAtoms in the program are deleted.
Here, our only \DLAtom is nonmonotonic because, assuming $p(t)$ is true, if $q(t)$ is true, the \DLAtom is false, but true if $q(t)$ is false.
Due to this nonmonotonicity, the \DLAtom is deleted and the programs's reduct is $p(t) \leftarrow$.
Eiter et al.'s strong semantics has two answer sets: The set that assigns $p(t)$ alone to be true ($\{ p(t) \}$) and the set that assigns
everything to be false ($\emptyset$).
Shen~\cite{dlwellsupported} recognizes that $p(t)$ is true due to circular justification and refines the
semantics by introducing a stronger entailment relation over \Interpretation ranges.
Under the well-supported semantics, for $p(t)$ to be true, it must be derivable from $\emptyset$. Because it is not, only $\emptyset$ is a well-supported
answer set.
\end{exampleOf}
The example above demonstrates an issue that arises due to deleting entire \DLAtoms when they are nonmonotonic.
There is a question as to whether the semantics could be defined differently so that a different definition of nonmonotonicity prevents cyclic dependencies.
However, Wang et al.~\cite{eliminatingnonmonotonicdlatoms} show that determining whether a \DLAtom is nonmonotonic in
general is intractable.
Another question is whether a well-supported semantics can be obtained by using a more granular reduct transformation.
Eiter et al.'s reduct transformation deletes entire \DLAtoms, but what if we were to delete individual operators instead?
It is known that without the $\intersectMinus$ operator, positive \DLAtoms are monotonic, thus a new reduct transformation could focus
on these operations alone.
Later on, we introduce a new semantics which leverages a more granular reduct transformation.

\section{Bound Semantics}\label{section:bound}

\subsection{Motivation}\label{section:why}
Anti\'{c} et al.~\cite{hexaft} show that the HEX AFT semantics (Approximation Fixpoint Theory) is equivalent to Shen's well-supported semantics for \DLPrograms (Theorem 8 in~\cite{hexaft}) and determining whether a program is consistent is $\Sigma_2^P$-complete.
That is, recognizing that a \DLProgram has a well-supported answer set can be done in polynomial time using a nondeterministic turing machine with access to an NP oracle.
In Eiter et al.'s original semantics for \DLPrograms, \DLAtoms are evaluated against a single \Interpretation, whereas the well-supported semantics requires a \DLAtom to be evaluated against an entire range of \Interpretations.
This range check is clearly the source of the increased complexity as it is coNP-hard in general~\cite{hexaft}.

In this section, we explore cases where \DLAtoms can be evaluated more efficiently.
First, observe that if a positive \DLAtom does not use the $\intersectMinus$ operator, that is, it only uses $\unionPlus$ and $\unionMinus$, then it is $\subseteq$-\Monotone w.r.t.\ \AnInterpretation.
For example, with $\phi \define DL[S \unionPlus p; S](t)$ we have $\emptyset \not\dlModels \phi$ and $\{ p(t) \} \dlModels \phi$.
As \AnInterpretation grows, the likelihood of a $\union$-\DLAtom being true increases.
For the $\intersectMinus$ operator, the monotonicity relation is reversed (i.e., it is \Antimonotone).
For example, with $\phi \define DL[S \intersectMinus p; \neg S](t)$ we have $\emptyset \dlModels \phi$ and $\{ p(t) \} \not\dlModels \phi$.
Unless a \DLAtom utilizes both types of operators with the same predicate, it can be evaluated efficiently.
We formally define a syntactic subclass of programs to show this property.

\begin{definitionOfWtitle}{dlatomaligned}{Aligned DL-Atoms}
    \ADLAtomO $\phi$ is {\em {\AlignedDLAtom}} if $\intersectMinus p \in \phi$ implies $\unionPlus p\not\in \phi$ and $ \unionMinus p \not\in \phi$.
    Otherwise, the atom is {\em \UnalignedDLAtom}.
    A \DLProgram that does not contain \UnalignedDLAtom \DLAtoms is \AlignedDLAtom.
\end{definitionOfWtitle}
For example, the \DLAtom $DL[S \unionPlus p, S' \intersectMinus p; S](t)$ is \UnalignedDLAtom because $p$ occurs in both types of operations, while $DL[S \unionPlus p, S' \unionMinus p; S](t)$ is \AlignedDLAtom.
Note that both types of operators can appear in \AnAlignedDLAtom \DLAtom if the operators use different predicates, e.g.\ $DL[S \unionMinus p, S \intersectMinus q; S](t)$ is \AlignedDLAtom.
We intend to show that the well-supported semantics can be simplified for \AlignedDLAtom \DLPrograms.

\AlignedDLAtomO \DLAtoms have an important property with regard to the $(E, I) \dlModels \phi$ entailment relation.
Namely, we can evaluate its contained $\union$-operators against one \Interpretation $E$ in the pair, and the $\intersect$-operators against the other \Interpretation $I$ rather than checking the entailment relation against every \Interpretation in the range $(E, I)$.
This shortcut is demonstrated in the following example.
\begin{exampleOf}{alignedshortcut}
    The \DLProgram $\Prog$ below  has no \StrongWellSupportedModel.
    \begin{align*}
        q(t) \leftarrow DL[S \unionPlus p, S' \intersectMinus q; S \sqcup \neg S'](t) \hspace{3cm}
        p(t) \leftarrow q(t)
    \end{align*}
    The set $\{ p(t), q(t) \}$ is not a well-supported answer set, thus $\lfp~\Gamma_{\Prog}(\cdot, \{ p(t), q(t) \})$ will not compute $\{ p(t), q(t) \}$.
    While computing $\lfp~\Gamma_{\Prog}(\cdot, \{ p(t), q(t) \})$, we check $\dlModels$ against $\emptyset$, $\{ p(t) \}$, $\{ q(t) \}$, and $\{ p(t), q(t) \}$.
    Because the \DLAtom is aligned, it is sufficient to only evaluate $\intersectMinus$ against $\{ p(t), q(t) \}$ (which has the fewest consequences considering $\intersectMinus$) and $\unionPlus$ against $\emptyset$ (which has the fewest consequences considering $\unionPlus$).
    As we will see, this shortcut lowers the complexity of the entailment relation $\dlModels$.
\end{exampleOf}

Next, we show \AnUnalignedDLAtom \DLProgram where checking every \Interpretation in a range is necessary under the well-supported semantics and cannot be shortcutted.

\begin{exampleOf}{alignedmissing}
    Let $\phi$ be the \DLAtom $DL[S \unionPlus p, S' \intersectMinus p; S \sqcup \neg S'](t)$ in the \DLProgram $\{ p(t) \leftarrow \phi \}$.
    Due to the overlapping use of the predicate $p$ between a $\union$- and $\intersect$-operator, $query_{\phi}$ behaves nonmonotonically.
    For the \Interpretation $\{ p(t) \}$, we have $query^{\union}_{\phi}(\{ p(t) \}) = \{ S(t) \}$, whereas for the \Interpretation $\emptyset$, which assigns $p(t)$ to be false, we have $query^{\intersect}_{\phi}(\emptyset) = \{ \neg S'(t) \}$.
    Both $query^{\union}_{\phi}(\emptyset)$ and $query^{\intersect}_{\phi}(\{ p(t) \})$ are equal to $\emptyset$.
    The \DLAtom is true under both \Interpretations $\{ p(t) \}$ and $\emptyset$.

    Let us consider the range $(E, I) = (\emptyset, \{ p(t) \})$.
    The relation $\dlModels$ must be checked against each \Interpretation in $(E, I)$ to determine whether $\{ p(t) \}$ is a \StrongWellSupportedModel of the program.
    Because both $query^{\union}_{\phi}(\emptyset)$ and $query^{\intersect}_{\phi}(\{ a \})$ are equal to $\emptyset$, we have $query^{\union}_{\phi}(E) \union query^{\intersect}_{\phi}(I) = \emptyset$.
    Thus, we cannot use the shortcut method described in \Example{alignedshortcut} to simplify the computation of $(E, I) \dlModels \phi$.
\end{exampleOf}

We intend to show that the shortcut method in \Example{alignedshortcut} can be applied in general to \AlignedDLAtom programs.
First, we introduce a new entailment relation $\dlModelsMin$ and later demonstrate that it is equivalent to $\dlModels$ on \AlignedDLAtom programs.
This new entailment relation also serves as the basis for our new semantics.
Rather than checking the ontology's entailment relation against each \Interpretation in a range, the new relation performs a single lower bound query (or upper bound for the case of negated \DLAtoms).

\begin{definitionOf}{dlmodelsmin}
    Let $\phi \define DL[S_i \mathbin{op_i} a_i, \dots; Q](\bm{c})$ and let $(E, I)$ be \AnInterpretation range.
\begin{align*}
    \definitionBody{dlqueries}{ queries_{\phi}(E, I) &\define \{ query_{\phi}(F) ~|~ E \subseteq F \subseteq I  \}} \\
    (E, I) \dlModelsMin \phi \hspace{0.35cm}&\textrm{ iff }\hspace{0.35cm} \intersectBig queries_{\phi}(E, I) \union \Ont \models Q(\bm{c}) \\
    (E, I) \dlModelsMin \Not~ \phi \hspace{0.35cm}&\textrm{ iff }\hspace{0.35cm} \unionBig queries_{\phi}(E, I) \union \Ont \not\models Q(\bm{c})
\end{align*}
Additional cases for $\dlModelsMin$ (e.g.\ for rules and atoms) are defined as in \Definition{dlmodels} by replacing all occurences of $\dlModels$ with $\dlModelsMin$.
We define $F \dlModelsMin \phi$ as shorthand for $(F, F) \dlModelsMin \phi$.
\end{definitionOf}

For a range of \Interpretations that only contains a single \Interpretation (e.g.\ $(I, I)$),
we can simplify both $\dlModelsMin$ and $\dlModels$ s.t.\ they are both equivalent to $query_{\phi}(I) \union \Ont \models Q(\bm{c})$.
\begin{lemma}\label{lemma:exactequivalence}
    We have $(I, I) \dlModelsMin \phi$ iff $(I, I) \dlModels \phi$.
\end{lemma}

The relation $\dlModelsMin$ checks a bound on all elements in $queries_{\phi}(E, I)$ instead of checking all elements inside this set.
We intend to formulate our new semantics around this relation and will show that these bounds can be computed efficiently.
We will also show that $\dlModelsMin$ and $\dlModels$ are equivalent for \AlignedDLAtom programs.
First, we return to \Example{alignedmissing} to demonstrate a case where $\dlModelsMin$ differs from $\dlModels$.

\begin{exampleOf}{ultimate}
    Let $\Prog$ be the rule $p(t) \leftarrow DL[S \unionPlus p, S' \intersectMinus p; S \sqcup \neg S'](t)$.
    Under Shen's semantics, $\Prog$ has a single answer set $\{ p(t) \}$.
    We have $(\emptyset, \{ p(t) \}) \dlModels (p(t) \leftarrow \phi)$ because $query_{\phi}(\emptyset) = \{ \neg S'(t) \}$, and $\neg S'(t) \models S(t) \sqcup \neg S'(t)$ and $query_{\phi}(\{ p(t) \}) = S(t)$, and we have $S(t) \models S(t) \sqcup \neg S'(t)$.
    For $\dlModelsMin$, we have $\intersectBig queries_{\phi}(\emptyset, \{ p(t)\}) = query_{\phi}(\emptyset) \intersect query_{\phi}(\{ p(t) \}) = \{ \neg S'(t) \} \intersect \{ S(t) \} = \emptyset$.
    Thus, $(\emptyset, \{ p(t) \}) \not\dlModelsMin \phi$.
\end{exampleOf}

The above establishes a difference between $\dlModels$ and $\dlModelsMin$.
The relation $\dlModels$ handles the presence of  $S \unionMinus q$ and $S \intersectMinus q$ in a \DLAtom $\phi$ by always including $\neg S(t)$ in $query_{\phi}(F)$ (assuming $q(t)$ is in the Herbrand base) regardless of the \Interpretation $F$.
The $\intersectMinus$ operator utilizes negation as failure, but if we create an analogous program using logic programming's negation as failure, then the analogy breaks down.
\begin{align*}
    q \leftarrow q \hspace{3cm}
    q \leftarrow \Not~q
\end{align*}
This program has no answer set under the stable model semantics~\cite{answersetsemantics}.
This program's Clark completion~\cite{completion} is given by the formula $q \iff q \lor \neg q$ which, under the law of the excluded middle, results in a tautology ($q \lor \neg q$) and $q$ must be true.
However, the completion semantics for normal logic programs is not considered to be well-supported~\cite{fagesws}.
Thus, we argue that a well-supported semantics for \DLPrograms should not treat $S \unionMinus q$ and $S \intersectMinus q$ in a manner analogous to a $q \lor \neg q$ by always including $S(t)$ in $query_{\phi}(F)$ for each $q(t)$ in the herbrand base.
This is the case for the $\dlModelsMin$ relation, but not for $\dlModels$ as demonstrated in \Example{ultimate}.
This argument provides some additional motivation for a new semantics formulated using the $\dlModelsMin$ relation.

We've established motivation and the foundations for a new semantics.
Namely, for the current well-supported semantics the consistency problem is $\Sigma_2^P$-complete, but it appears that a large class of programs (\AlignedDLAtom programs) can be checked in NP-time.
Additionally, the current well-supported semantics share properties with a program completion, which is not well-supported.
We advance to define our semantics and some technical properties.

\subsection{A Fixpoint Characterization}\label{section:new}

We construct a new immediate consequence operator by replacing the entailment relation $\dlModels$ in Shen's semantics (\Definition{dlmodelsconsequence}) with the new relation $\dlModelsMin$ defined in the previous section (\Definition{dlmodelsmin}).
\begin{definitionOf}{minSWSS}
$\Gamma^{\wedge}_{\Prog}(E, I) \define \{ \head(r) ~|~ r \in \Prog, (E, I) \dlModelsMin \body(r) \}$.
\end{definitionOf}

We replace $\Gamma_{\Prog}$ with $\Gamma^{\wedge}_{\Prog}$ in the definition of well-supported answer sets (\Definition{strongwellsupportedmodel}).
\begin{definitionOf}{fixpointalignedanswerset}
    \AnInterpretationO $I$ is a {\em \AlignedAnswerSet} of a \DLProgram $\Prog$ if $I = \lfp~\Gamma^{\wedge}_{\Prog}(\cdot, I)$.
\end{definitionOf}

This establishes our new semantics.
Before we examine its properties, we must show that it is well-defined.
Namely, that a least fixpoint exists.
\definitionBody{monotone}{A function $o$ is {\em \Monotone} (resp.\ {\em \Antimonotone}) w.r.t.\ an ordering $\prec$ if $x \prec y$ implies $o(x) \prec o(y)$ (resp.\ $o(y) \prec o(x)$).}
\begin{lemma}\label{lemma:querymonotone}
    Given a \DLAtom $\phi$, the function $query_{\phi}^{\union}(\cdot)$ is $\subseteq$-\Monotone  and $query_{\phi}^{\intersect}(\cdot)$ is $\supseteq$-\Antimonotone.
\end{lemma}
Next, we show that the entire entailment relation $\dlModelsMin$ is \Monotone by adopting the $\ltePrecision$ ordering from three- and four-valued logics~\cite{belnapfour} s.t.\
$(E, I) \ltePrecision (E', I')$ if $E \subseteq E'$ and $I \supseteq I'$.
\begin{lemma}\label{lemma:dlalignedentailmentmonotone}
    If $(E, I) \dlModelsMin \phi$ then $(E', I') \dlModelsMin \phi$ where $(E, I) \ltePrecision (E', I')$.
\end{lemma}

This monotonicity relation is tighter than the one shown by Shen~\cite{dlwellsupported}, which only looks at the first component of pairs.
We obtain the analog to Shen's result immediately following \Lemma{dlalignedentailmentmonotone} above.
\begin{corollaryOf}{dlalignedentailmentmonotonesimple}
    If $(E, I) \dlModelsMin \phi$ then $(F, I) \dlModelsMin \phi$ where $E \subseteq F \subseteq I$.
\end{corollaryOf}
Due to \Lemma{dlalignedentailmentmonotone}, the $\Gamma^{\wedge}_{\Prog}$ operator is monotone w.r.t.\ $\ltePrecision$ and $\subseteq$.
\begin{lemma}
    Given $(E, I) \ltePrecision (E', I')$, we have $\Gamma^{\wedge}_{\Prog}(E, I) \subseteq \Gamma^{\wedge}_{\Prog}(E', I')$.
\end{lemma}
Because $\Gamma_{\Prog}^{\wedge}(\cdot, I)$ is $\subseteq$-\Monotone, a least fixpoint exists~\cite{tarskilatticetheoretical1955} and \Definition{fixpointalignedanswerset} is well-defined.
Immediately, because the relations $\dlModelsMin$ and $\dlModels$ are equivalent for a single \Interpretation (\Lemma{exactequivalence}), their corresponding operators are equivalent for a single \Interpretation.
\begin{corollaryOf}{exactequivalence}
    $\Gamma_{\Prog}(I, I) = \Gamma^{\wedge}_{\Prog}(I, I)$.
\end{corollaryOf}

We now further relate $\dlModelsMin$ and $\dlModels$ so that we can compare $\Gamma^{\wedge}_\Prog$ with $\Gamma_{\Prog}$.
\begin{proposition}\label{proposition:entailmenteq}
    We have
    $(E, I) \dlModels \omega$ if $(E, I) \dlModelsMin \omega$
    where $E \subseteq I$. The relation is ``iff'' all \DLAtoms contained in $\omega$ are \AlignedDLAtom.
\end{proposition}

As a relation, $\dlModelsMin$ is a subset of $\dlModels$.
Further, the relations are equivalent for \AlignedDLAtom programs.
Intuitively, this is because there is an \Interpretation in the range $(E, I)$ which, when evaluated, is equivalent to evaluating the bounds.
We demonstrate this property formally in the following.
\begin{proposition}\label{proposition:existsgeneratinginterpretation}
    For any  \AlignedDLAtom \DLAtom $\phi$ and $E \subseteq I$,
    there exist $S^+$ and $S^-$ s.t.
    $E \subseteq S^+ \subseteq I$ and $E \subseteq S^- \subseteq I$ and
    \begin{align*}
        query_{\phi}(S^+) &= \intersectBig queries_{\phi}(E, I) \\
        query_{\phi}(S^-) &= \unionBig queries_{\phi}(E, I)\\
        S^+ &\define E \union \{ p(t) \in I ~|~ \textrm{if $\intersectMinus\hspace{2px} p \in \phi$} \}\\
        S^- &\define E \union \{ p(t) \in I ~|~ \textrm{if $\unionPlus\hspace{2px} p \in \phi$ or $\unionMinus\hspace{2px} p \in \phi$} \}
    \end{align*}
\end{proposition}

For positive \DLAtoms, the \Interpretation $E$ contains all atoms that are true in every \Interpretation in the range, while the complement of $I$ contains all atoms that are never true in the range.
Thus, to compute the lower bound of all queries ($S^+$) we extend $E$ with the predicates that will have an effect on an $\intersectMinus$-operator.
Due to the \AlignedDLAtom property, predicates in an $\intersectMinus$-operator will not appear in a $\union$-operator in the \DLAtom.
That is, the only atoms in $S^+$ that will activate a $\union$-operator also appear in $E$.
For negated \DLAtoms ($S^-$), the $\union$-operators are nonmonotonic and $\intersect$-operators are monotone (we analyze this property in further detail in \Section{reduct}), thus $S^-$ is constructed using the $\union$-operators instead of $\intersect$.

In \Example{ultimate}, we demonstrated a case where $\dlModels$ and $\dlModelsMin$ differ.
We can begin to explain this phenomenon using \Proposition{existsgeneratinginterpretation}.
The intent behind the construction of $S^+$ is that $E$ contains the lower bound for all $\union$-operators and $\{ p(t) \in I ~|~ \textrm{if $\intersectMinus p \in \phi$} \}$ is the lower bound for $\intersect$-operators.
However, for \UnalignedDLAtom \DLAtoms, the ideal lower bound is not two-valued.
We may require that an atom be not true and not false.
Thus, we cannot use a single \Interpretation to simulate $\dlModelsMin$ for \UnalignedDLAtom \DLAtoms.
Instead, we construct an alternative query which allows us to have a claim similar to \Proposition{existsgeneratinginterpretation} that works for \UnalignedDLAtom programs.

\begin{proposition}\label{proposition:unalignedexistsgeneratinginterpretation}
    For any (possibly \UnalignedDLAtom) \DLAtom $\phi$, and $E \subseteq I$
    we have
    \begin{align*}
        \intersectBig queries_{\phi}(E, I) &=
            query_{\phi}^{\union}(E) \union query_{\phi}^{\intersect}(I) \union \{ \neg S(\bm{e}) ~|~ (S \unionMinus p), (S \intersectMinus p) \in \phi, p(\bm{e}) \in HB \} \\
        \unionBig queries_{\phi}(E, I) &= query_{\phi}^{\union}(I) \union query_{\phi}^{\intersect}(E)
    \end{align*}
\end{proposition}

It is straightforward to show that \Proposition{existsgeneratinginterpretation} follows from the above when we restrict to \AlignedDLAtom \DLAtoms because there is no overlap between $query_{\phi}^{\union}(E)$ and $query_{\phi}^{\intersect}(I)$.
At last, we have formalized the shortcut demonstrated in \Example{alignedshortcut}.
Similar to how the binary $\Gamma_{\Prog}(E, I)$ operator uses $E$ to evaluate positive atoms and $I$ to evaluate $\Not$ atoms in a program,
\Proposition{unalignedexistsgeneratinginterpretation} shows that we can use $E$ to evaluate one half of the query and $I$ to evaluate the other half.

It is clear by \Proposition{unalignedexistsgeneratinginterpretation} that we can construct $\unionBig queries_{\phi}(E, I)$ or $\intersectBig queries_{\phi}(E, I)$ in linear time, thus $\dlModelsMin$ is tractable, unlike $\dlModels$.
Thus, the \AlignedAnswerSet semantics are NP-complete while the well-supported semantics are $\Sigma_2^P$-complete~\cite{hexaft}.

We lift the relationship between $\dlModelsMin$ and $\dlModels$ established in \Proposition{entailmenteq} to operators.
\begin{corollaryOf}{entailementsubseteq}
    Given $E \subseteq I$, we have
    $\Gamma^{\wedge}_{\Prog}(E, I) \subseteq \Gamma_{\Prog}(E, I)$ and if $\Prog$ is \AlignedDLAtom, then $\Gamma^{\wedge}_{\Prog}(E, I) = \Gamma_{\Prog}(E, I)$.
\end{corollaryOf}
The well-supported and \AlignedAnswerSet semantics are equivalent for \AlignedDLAtom programs.
Because the operators are the same for a single \Interpretation (\Corollary{exactequivalence}) and due to the subset relation outlined above (\Corollary{entailementsubseteq}), we can make a powerful connection between least fixpoints.
\begin{proposition}\label{proposition:bigconnection}
    If $\lfp~\Gamma_{\Prog}^{\wedge}(\cdot, I) = I$ then $\lfp~\Gamma_{\Prog}(\cdot, I) = I$.
\end{proposition}
% \begin{proofE}
%     (By contradiction)
%     Assume $\lfp~\Gamma_{\Prog}^{\wedge}(\cdot, I) = I$ and $\lfp~\Gamma_{\Prog}(\cdot, I) \not= I$.
%     We have $\Gamma_{\Prog}^{\wedge}(I, I) = I$, thus $\Gamma_{\Prog}(I, I) = I$ (\Corollary{exactequivalence}).
%     Because $I$ is a fixpoint of $\Gamma_{\Prog}(\cdot, I)$, it follows that
%     $\lfp~\Gamma_{\Prog}(\cdot, I) \subset I$.
%     Let $W \define \lfp~\Gamma_{\Prog}(\cdot, I)$.
%     By \Corollary{entailementsubseteq}, we have $\Gamma^{\wedge}_{\Prog}(W, I) \subseteq \Gamma_{\Prog}(W, I)$.
%     With $\Gamma_{\Prog}(W, I) = W \subset I$ and by the transitivity of $\subseteq$, we have $\Gamma_{\Prog}^{\wedge}(W, I) \subset I$.
%     That is, $W$ is a prefixpoint of $\Gamma_{\Prog}^{\wedge}(\cdot, I)$ which violates the assumption that $I$ is the least fixpoint of $\Gamma_{\Prog}^{\wedge}(\cdot, I)$ because the least fixpoint is also the least prefixpoint~\cite{tarskilatticetheoretical1955}.
% \end{proofE}

Because both semantics are defined in terms of least fixpoints (Definitions \ref{definition:strongwellsupportedmodel} and \ref{definition:fixpointalignedanswerset}), we establish that our new semantics is a strict subset of the well-supported semantics following directly from  \Proposition{bigconnection}.
\begin{theorem}
    Every \AlignedAnswerSet of $\Prog$ is a \WellSupportedSemantics answer set of $\Prog$.
\end{theorem}

Thus, our new semantics is a strict subset of Shen's well-supported semantics, and is equivalent for \AlignedDLAtom \DLPrograms.
Shen\ introduces well-supported semantics using a well-founded ordering.
For brevity, we have skipped this step, however, it is straightforward to construct a well-founded ordering using iterations of the $\Gamma^{\wedge}_{\Prog}$ operator.
For example, $\Gamma^{\wedge}_{\Prog}(\emptyset, I) < \Gamma^{\wedge}_{\Prog}(\Gamma^{\wedge}_{\Prog}(\emptyset, I), I) < \dots$.
Due to the subset relationship between semantics, our new semantics inherits Shen's well-supportedness property.

In logic programming, reduct-based characterizations are common.
This is how Eiter et al.'s original semantics for \DLPrograms was presented.
In the sequel, we present an alternative characterization of the \AlignedAnswerSet semantics (\Definition{fixpointalignedanswerset}) using a reduct transformation.
This characterization provides additional insights into its fixpoint representation and enables us to further simplify our fixpoint semantics.
\subsection{A Reduct-Based Characterization}\label{section:reduct}

The reduct-based semantics of logic programs~\cite{answersetsemantics} uses an \Interpretation to evaluate and remove all nonmonotonic portions of a program.
The resulting program, called the reduct, is a propositional program and thus it has a unique minimal model.
\AnAnswerSetO is \AnInterpretation that is both a model of a program and the unique minimal model of the program's reduct w.r.t.\ the \Interpretation.
This is also how Eiter et al.\ initially characterized their semantics for \DLPrograms.

In this section, we characterize our new \AlignedAnswerSet semantics using a reduct-based approach.
We gain additional insight into \DLPrograms, namely, we define a new syntactic subset of \DLPrograms analogous to propositional logic programs.
Unlike propositional programs, our definition permits some negated \DLAtoms.

First, we define a model in terms of $\dlModelsMin$.
\begin{definitionOf}{alignedmodel}
    \AnInterpretationO $I$ is {\em \AnAlignedModel model} of a \DLProgram $\Prog$ if for each $r \in \Prog$, $I \dlModelsMin r$.
\end{definitionOf}

We observe something surprising for well-supported and \AlignedModel answer set semantics:
a large syntactic class of negated \DLAtoms are monotone.
In positive \DLAtoms, the $\intersectMinus$ operator is nonmonotonic while the $\unionPlus$/$\unionMinus$ operators are monotonic.
For \DLAtoms that appear in a $\Not$, the roles of these operators is reversed.
We demonstrate this concretely in the following example.
\begin{example}
    Let $\phi$ be the \DLAtom $DL[S \unionPlus p; S](t)$.
    Here, we have $\emptyset \dlModelsMin \Not~ \phi$ and $\{ p(t) \} \not\dlModelsMin \Not~ \phi$.
    Introducing a new atom, $p(t)$ decreases the consequences of $\dlModelsMin$, that is, $\unionPlus$ behaves nonmonotonically when inside a $\Not$.
    If we use $\phi' \define DL[S \intersectMinus p; S](t)$, then we have the opposite.
    That is, $\emptyset \not\dlModelsMin~ \Not~ \phi$ and $\{ p(t) \} \dlModelsMin \Not~ \phi$.
    The $\intersectMinus$ operator behaves monotonically when inside a $\Not$.
\end{example}

Because negative $\intersect$-\DLAtoms are \Monotone, they should not be removed from a program when constructing its reduct.

With the reduct operation for logic programs, negative atoms are removed from the body of rules as a method of partial evaluation.
However, removing nonmonotonic operators from a \DLAtom alters the meaning of the \DLAtom.
Rather than removing operators, we want to partially evaluate them.
For example, if we want to effectively partially evaluate $\phi \define DL[S \unionPlus p; Q](t)$ with the \Interpretation $\{ p(r) \}$, we can introduce a new predicate $p'$, replace $S \unionPlus p$ with $S \unionPlus p'$,  and add a rule $p'(r) \leftarrow$ to our program.
This will ensure that (for \Interpretations that satisfy all rules) $S(r)$ will be a part of the query evaluation.

Instead of performing this transformation, we introduce a new operator $\unionTop$ as syntactic sugar.
The operator appears alongside other operators in the form $S \unionTop Y$ or $\neg S \unionTop Y$ where $S$ is a role or concept and $Y$ is a set of atoms.
We modify the evaluation of $query_{\phi}(I)$ by adding additional elements due to this new operator.
\begin{align*}
S(\bm{e}) \in query_{\phi}(I) \textrm{ if } (S \unionTop Y) \in \phi \textrm{ and } p(\bm{e}) \in Y \hspace{1cm}
\neg  S(\bm{e}) \in query_{\phi}(I) \textrm{ if } (\neg S \unionTop Y) \in \phi \textrm{ and } p(\bm{e}) \in Y
\end{align*}
Recall that $\union$-\DLAtoms do not contain $\intersectMinus$ and $\intersect$-\DLAtoms do not contain $\unionPlus$ or $\unionMinus$.
We extend the definitions of $\union$- and $\intersect$-\DLAtoms to also permit the $\unionTop$ operator.
\begin{remark}
    Earlier, we noted that Shen's well-supported semantics treats the presence of $S \unionMinus q$ and $S \intersectMinus q$ in the same \DLAtom by always adding $\neg S(t)$ to $query_{\phi}(I)$ regardless of $I$. Another way of framing this property is that, given an \Interpretation $I$, both operators can be replaced with $\neg S \unionTop \{ q(\bm{e}) ~|~ q(\bm{e}) \in I \}$.
\end{remark}
We introduce a syntactic subset of \DLPrograms to serve as an analog to propositional logic programs.

\begin{definitionOf}{propositionaldl}
    \ADLProgramO $\Prog$ is {\em \AlignedPositive} if, every positive (resp. negative) \DLAtom is a $\union$-\DLAtom (resp.\ an $\intersect$-\DLAtom) and $\Not~p$, where $p$ is an atom, does not occur in the program.
\end{definitionOf}
\begin{remark}
    An \AlignedPositive \DLProgram $\Prog$ is \AlignedDLAtom.
\end{remark}

Operations involving $\unionTop$ are monotonic because $\unionTop$ is a constant operation, it is the same regardless of the \Interpretation.
The $\union$-operators are nonmonotonic in negative \DLAtoms while the $\intersect$-operator is nonmonotonic for positive \DLAtoms.
Thus, we call a \DLAtom \AlignedPositive if it does not have such \DLAtoms or standard negation as failure applied to plain atoms.
If we remove all \DLAtoms from an \AlignedPositive \DLProgram, the resulting program is propositional.
Similar to answer set programming and Eiter et al.'s semantics for \DLPrograms,
we intend to define a reduct transformation that converts any \DLProgram to an \AlignedPositive one given \AnInterpretation.
This transformation will enable us to characterize our new semantics as the minimal models of these \AlignedPositive programs.
Before we introduce this transformation, we observe the following.

\begin{lemma}\label{lemma:propositionalnoreduct}
    For an \AlignedPositive \DLProgram $\Prog$ and \Interpretation ranges $(E, I)$ and $(E, I')$, we have $\Gamma_{\Prog}(E, I) = \Gamma_{\Prog}(E, I')$.
\end{lemma}

Recall that $\Gamma_{\Prog}^{\wedge}(E, I)$ and $\Gamma_{\Prog}(E, I)$ are equivalent for \AlignedPositive \DLPrograms due to \Corollary{entailementsubseteq}.
The above shows that the second parameter of the $\Gamma_{\Prog}(\cdot, \cdot)$ operator has no impact on the computation if $\Prog$ is \AlignedPositive.
The second parameter is used to evaluate the nonmonotonic components of the program, the portion of the program that needs to be eliminated by a reduct transformation.

We can use this insight to simplify $\Gamma_{\Prog}$ into a unary operator.
We write $\Gamma_{\Prog}(E)$ to denote any invocaton of $\Gamma_{\Prog}(E, I)$ where $E \subseteq I$, all of which are equivalent due to \Lemma{propositionalnoreduct}.
Now, with a unary immediate consequence operator $\Gamma_{\Prog}(\cdot)$, we make the following claim.
\begin{proposition}\label{proposition:propositionaldlmodel}
    Every \AlignedPositive \DLProgram $\Prog$ has a unique $\subseteq$-minimal \AlignedModel model $\lfp~\Gamma_{\Prog}(\cdot)$.
\end{proposition}

Just like Eiter et al.'s \DLProgram semantics, our \AlignedPositive \DLProgram is guaranteed to have a unique minimal model.
Next, we define a reduct transformation for \DLPrograms that transforms an arbitrary program into an \AlignedPositive one based on an \Interpretation.
\begin{definitionOf}{aligningreduct}
    Given \ADLProgram $\Prog$ and \AnInterpretation $I$, the \AlignedReduct $\alignedreduct{\Prog}{I}$ is obtained by performing the transformations on each rule $r \in \Prog$.

    \begin{itemize}
        \item  For each $\Not~ q \in \body(r)$ (where $q$ is a plain atom), if $q \in I$, delete the rule, otherwise, remove $\Not~ q$ from the rule body.
    \item For each positive \DLAtom,
replace $S \intersectMinus p$ with $\neg S \unionTop \{ p(\bm{e}) ~|~ p(\bm{e}) \in HB \setminus I \}$
\item For each negative \DLAtom,
         replace $S \unionPlus p$ with $S \unionTop \{ p(\bm{e}) ~|~ p(\bm{e}) \in I \}$ and replace $S \unionMinus p$ with $\neg S \unionTop \{ p(\bm{e}) ~|~ p(\bm{e}) \in I \}$.
    \end{itemize}
\end{definitionOf}

We remove all $\union$-operators from negative \DLAtoms, all $\intersect$-operators from positive \DLAtoms, and all instances of $\Not~q$, so a program resulting from the reduct transformation is \AlignedPositive.
\begin{lemmaOf}{reductispropositional}
    Given \AnInterpretation $I$ and \DLProgram $\Prog$, the program $\alignedreduct{\Prog}{I}$ is \AlignedPositive.
\end{lemmaOf}

We now define a semantics that mirrors the reduct-based semantics for logic programs.
\begin{definitionOf}{alignedanswerset}
    \AnInterpretationO $I$ is {\em \AnAlignedAnswerSet} of $\Prog$ if
it is the $\subseteq$-minimal model of $\alignedreduct{\Prog}{I}$.
\end{definitionOf}
Because \AlignedPositive programs have a $\subseteq$-minimal \AlignedModel model (\Proposition{propositionaldlmodel}) and $\alignedreduct{\Prog}{I}$ is \AlignedPositive (\Lemma{reductispropositional}), we can characterize the $\subseteq$-minimal \AlignedModel model of $\alignedreduct{\Prog}{I}$ as $\lfp~\Gamma_{\alignedreduct{\Prog}{I}}(\cdot)$.

Looking at \Lemma{propositionalnoreduct}, we wonder whether the second parameter of the $ \Gamma_{\Prog}$ operator performs the same function of the reduct transformation.
We demonstrate that this is the case.

\begin{lemma}\label{lemma:upperboundisreduct}
   $\Gamma_{\alignedreduct{\Prog}{I}}(E) = \Gamma^{\wedge}_{\Prog}(E, I)$.
\end{lemma}

In \Proposition{propositionaldlmodel}, we established that \AlignedPositive programs have a unique minimal model that can be computed as the least fixpoint of $\Gamma_{\Prog}(\cdot)$.
Because our new well-supported semantics (\Definition{minSWSS}) are defined in terms of the binary operator $\Gamma^{\wedge}_{\Prog}$ and because the reduct performs the function of the second parameter of the operator (\Lemma{upperboundisreduct}), it is straightforward to equate our reduct-based semantics (\Definition{alignedanswerset}) with our fixpoint-based semantics (\Definition{fixpointalignedanswerset}).

\begin{theorem}
    Let $\Prog$ be \ADLProgram and $M$ \AnInterpretation.
    $M$ is \AnAlignedModel model of $\Prog$ and it is the $\subseteq$-minimal model of $\alignedreduct{\Prog}{M}$ (\Definition{alignedanswerset})
    iff $M = \lfp~\Gamma^{\wedge}_{\Prog}(\cdot, M)$
    (\Definition{fixpointalignedanswerset}).

\end{theorem}

 \section{Discussion}\label{section:discussion}
We have introduced a new well-supported semantics, the \AlignedAnswerSet semantics, for \DLPrograms and provided a fixpoint characterization and
a reduct-based characterization.
We have shown that the new semantics is stricter than Shen's well-supported semantics.
Namely, our entailment relation is stricter and every \AlignedAnswerSet is a well-supported answer set.
This has the consequence that every answer set under the new semantics is well-supported by Shen's definition of well-supportedness.
Our new semantics also induces a stricter notion of well-supportedness via its fixpoint characterization.
We showed that some well-supported answer sets missed by our semantics contain cyclical dependencies if compared to a program's Clark completion.

While this work focused exclusively on the strong well-supported semantics, Shen\ also defines a weak well-supported semantics where cyclic dependencies are permitted to exist across negated \DLAtoms.
We showed that a negated DL atom can be monotonic and thus is appropriate to include our analog to propositional programs: \AlignedPositive programs.
This suggests that Shen's strong variant of well-supported semantics is more appropriate than the weak variant as a $\Not$-negated $\intersect$-\DLAtom is monotonic but is removed by Shen's weak semantics.

We have identified the class of \AlignedDLAtom \DLPrograms for which our new semantics is equivalent to the well-supported semantics.
This class is analoguous to head-cycle free programs for disjunctive logic programming.
The complexity of determining whether \DLProgram has a well-supported answer set under Shen's semantics is NP-complete if the program is \AnAlignedDLAtom
(one can test whether it has \AnAlignedModel answer set), but it is $\Sigma_2^P$-complete in general.

Our reduct-based characterization of the \AlignedAnswerSet semantics identifies a useful class of \DLPrograms, namely \AlignedPositive \DLPrograms.
Programs in this class are monotonic and thus always have unique minimal model.
Because this subclass is \AlignedDLAtom, these minimal models can be computed using Shen's fixpoint operator or using our operator for the \AlignedAnswerSet semantics.
Nonmonotonic aspects of \DLPrograms create challenges in other research.
When Motik and Rosati~\cite{motikreconciling2010} define hybrid MKNF knowledge bases, another rule and ontology-based
formalism, they show that their formalism can express $\union$-\DLPrograms under Eiter et al.'s semantics.
They do not consider the $\intersectMinus$ operator, however, Eiter and Simkus~\cite{kbns} show that Hybrid MKNF can be encoded as a \DLProgram.
It remains an open question whether Shen's well-supported semantics or our new \AlignedModel semantics can be expressed
as a hybrid MKNF knowledge base.
Wang et al.~\cite{eliminatingnonmonotonicdlatoms} explore removing $\intersectMinus$ from \DLPrograms, and this research
could be extended to well-supported and \AlignedModel semantics.
Rather than removing $\intersectMinus$ from programs, it would also be interesting to convert programs into an \AlignedPositive programs s.t.\ Eiter et al.'s semantics is equivalent to the \AlignedAnswerSet semantics of the original program.

It is likely our operator can be generalized to a symmetric approximator under Approximation Fixpoint Theory (AFT)~\cite{denecker2000approximations}.
Such a generalization would induce a three-valued semantics and a well-founded semantics.
The AFT characterization for the well-supported semantics relies on consistent AFT~\cite{DeneckerMT04} which is more complex than symmetric AFT~\cite{denecker2000approximations}.

% \section{Proofs}\label{section:proofs}
% \printProofs

\bibliographystyle{eptcs}
\bibliography{references}

\end{document}